\documentclass[english]{article}
\usepackage[utf8]{inputenc}
\usepackage[T1]{fontenc}
\usepackage{babel}
\usepackage{amsmath}
\usepackage{graphicx}
\usepackage{fancyhdr}
\usepackage{lineno,hyperref}
\usepackage{multirow}
\newcommand{\romannum}[1][1]{\uppercase\expandafter{\romannumeral#1}}

\begin{document}

\title{A Remote Sensing Image Dataset for Cloud Removal}

\author{Daoyu Lin\textsuperscript{1}, Guangluan Xu\textsuperscript{1}, Xiaoke Wang\textsuperscript{1, 2}, \\Yang Wang\textsuperscript{1}, Xian Sun\textsuperscript{1}, Kun Fu\textsuperscript{1, 2}}

\maketitle
\thispagestyle{fancy}

1. Institute of Electronics, Chinese Academy of Sciences, Beijing, China \\2. Department of Electronic, Electrical and Communication Engineering,
University of Chinese Academy of Sciences, Beijing, China
\begin{abstract}
Cloud-based overlays are often present in optical remote sensing images, thus limiting the application of acquired data. Removing clouds is an indispensable pre-processing step in remote sensing image analysis. Deep learning has achieved great success in the field of remote sensing in recent years, including scene classification and change detection. However, deep learning is rarely applied in remote sensing image removal clouds. The reason is the lack of data sets for training neural networks. In order to solve this problem, this paper first proposed the Remote sensing Image Cloud rEmoving dataset (RICE). The proposed dataset consists of two parts: RICE1 contains 500 pairs of images, each pair has images with cloud and cloudless size of 512*512; RICE2 contains 450 sets of images, each set contains three 512*512 size images. , respectively, the reference picture without clouds, the picture of the cloud and the mask of its cloud. The dataset is freely available at \url{https://github.com/BUPTLdy/RICE_DATASET}.
\end{abstract}

\section{Introduction}

With the development of remote sensing technology, satellite imagery plays a very important role in various applications, such as Earth observation, climate change, and environmental monitoring. However, optical remote sensing images are often contaminated by clouds and clouds, and cloud occlusion is a serious problem in optical images. Both cloud and cloud shadows will reduce the utilization of image data and limit the use of these optical remote sensing images in further applications. Therefore, removing the cloud is necessary to improve the utilization of optical remote sensing images.

With the development of deep learning, convolutional neural networks have made great progress in low-level computer vision tasks, such as image completion, image defogging, image denoising, and image super-resolution. At the same time, deep learning has brought new developments in the field of remote sensing image research, such as remote sensing image segmentation and remote sensing image detection.

\begin{figure}[!htb]
	
	\hspace{10mm} \verb|Cloud|
	\hspace{18mm} \verb|Label|
	\hspace{18mm} \verb|Cloud|
	\hspace{18mm} \verb|Label|
	\vspace{-2mm}
	\begin{center}
		\includegraphics[width=0.24\textwidth]{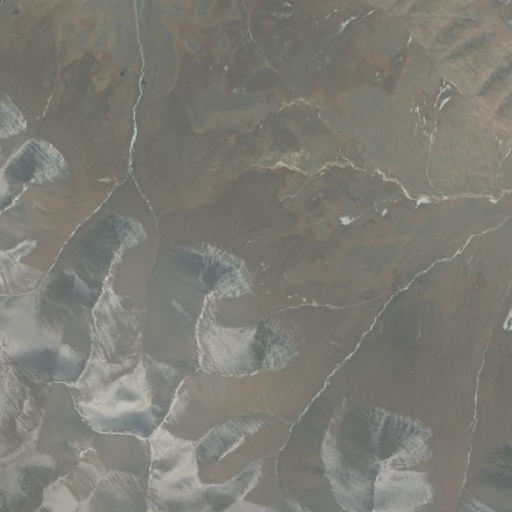}
		\includegraphics[width=0.24\textwidth]{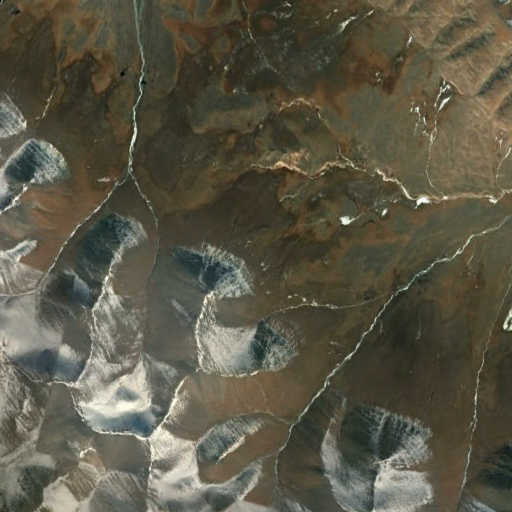}
		\includegraphics[width=0.24\textwidth]{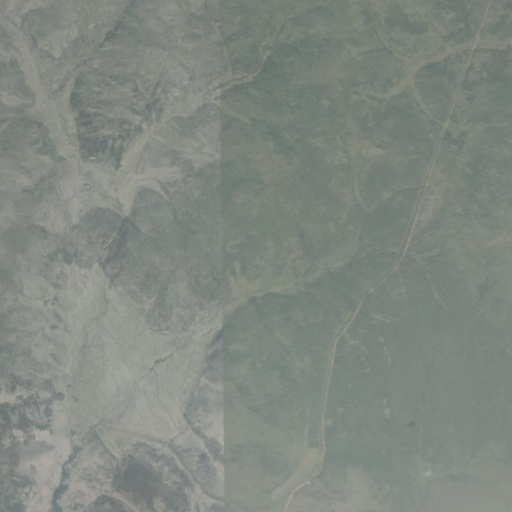} 
		\includegraphics[width=0.24\textwidth]{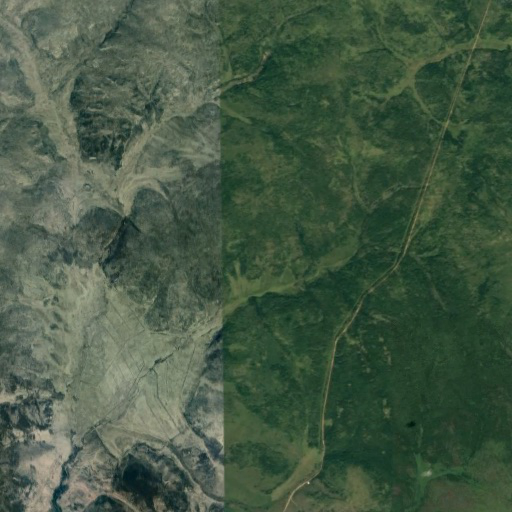} 
		\\
		\vspace{2mm}
		\includegraphics[width=0.24\textwidth]{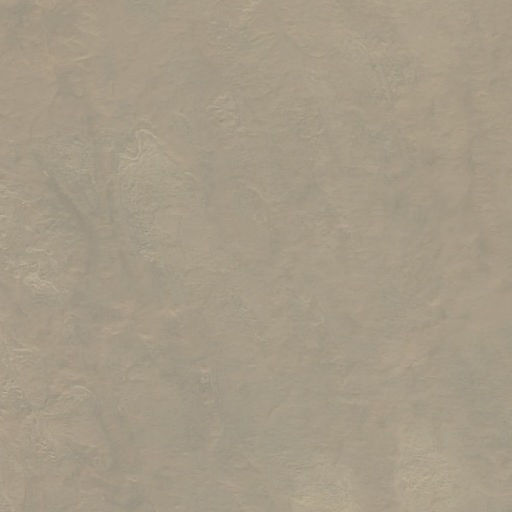}
		\includegraphics[width=0.24\textwidth]{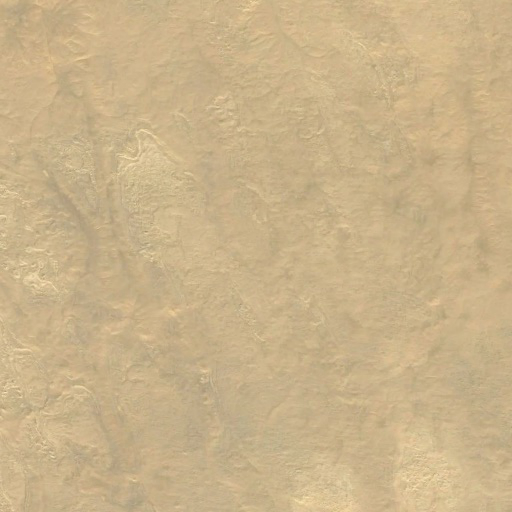}
		\includegraphics[width=0.24\textwidth]{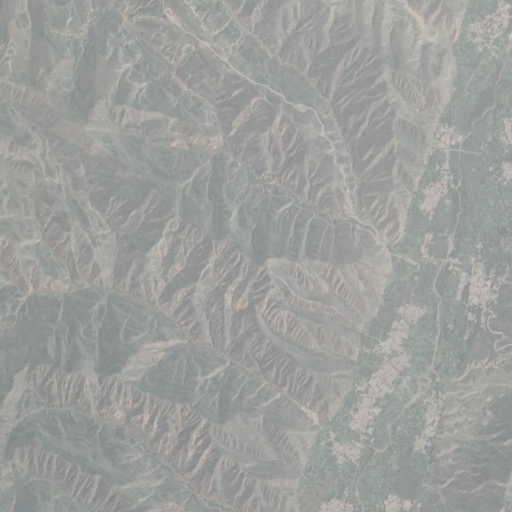} 
		\includegraphics[width=0.24\textwidth]{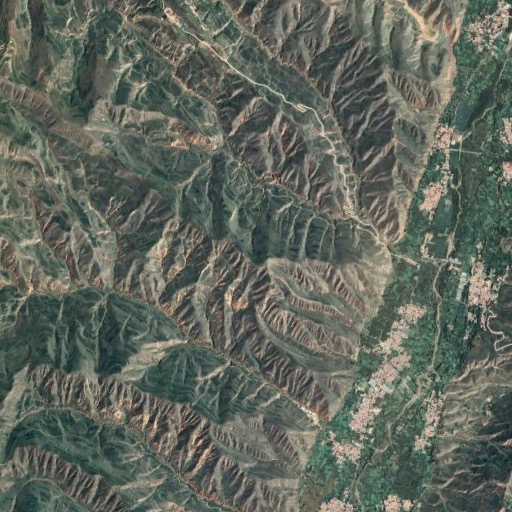} 
		\\
		\vspace{2mm}
		\includegraphics[width=0.24\textwidth]{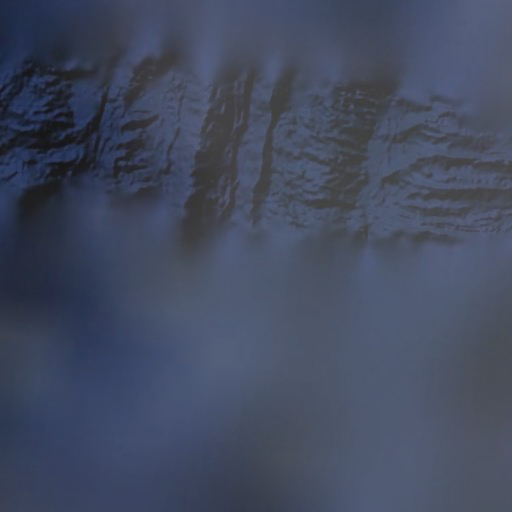}
		\includegraphics[width=0.24\textwidth]{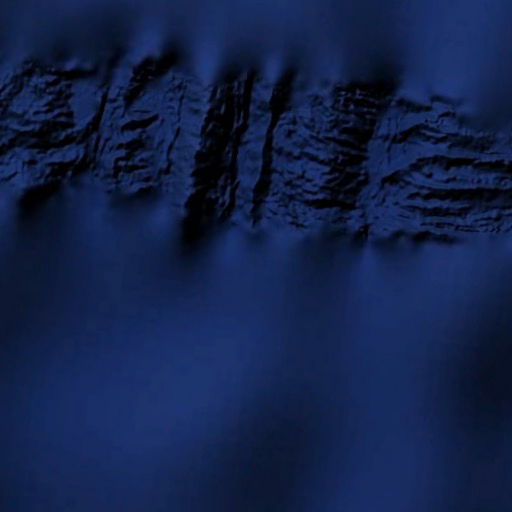}
		\includegraphics[width=0.24\textwidth]{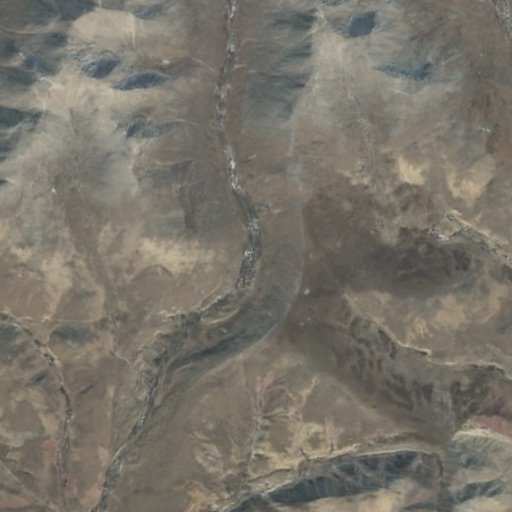} 
		\includegraphics[width=0.24\textwidth]{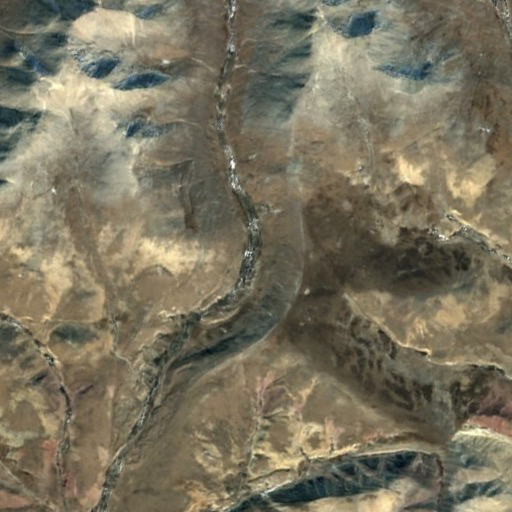} 
		\\
		\vspace{2mm}
		\includegraphics[width=0.24\textwidth]{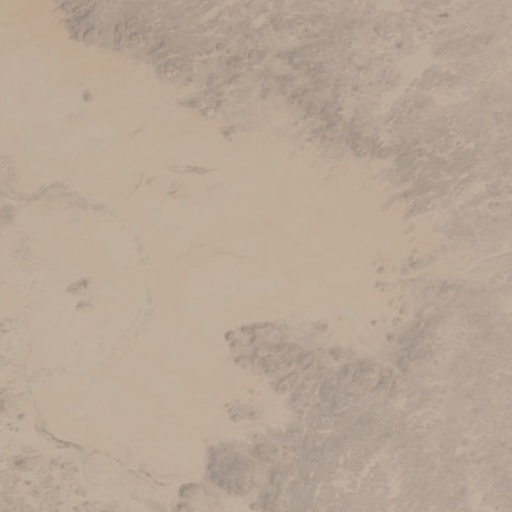}
		\includegraphics[width=0.24\textwidth]{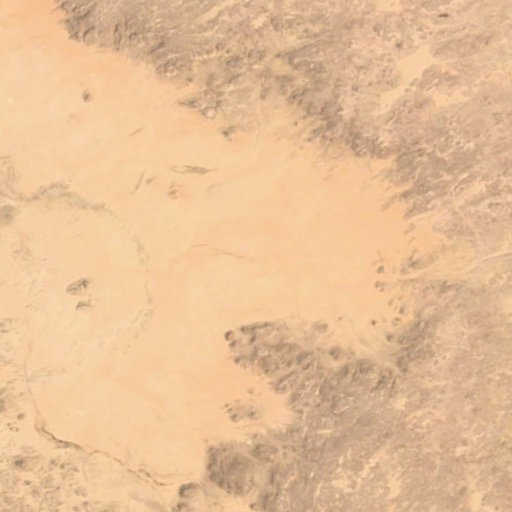}
		\includegraphics[width=0.24\textwidth]{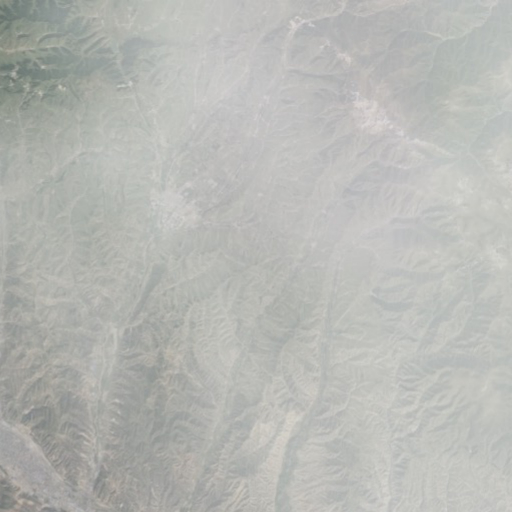} 
		\includegraphics[width=0.24\textwidth]{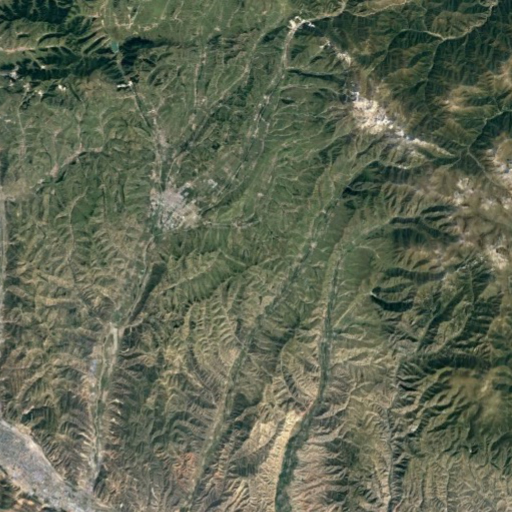} 
	\end{center}
	\vspace{-5mm}
	\caption{RICE-\romannum[1] dataset.}
	\label{fig:rice1}
\end{figure}

\begin{figure}[!htb]
	
	\hspace{25mm} \verb|Cloud|
	\hspace{18mm} \verb|Mask|
	\hspace{18mm} \verb|Label|
	\vspace{-2mm}
	\begin{center}
		\includegraphics[width=0.24\textwidth]{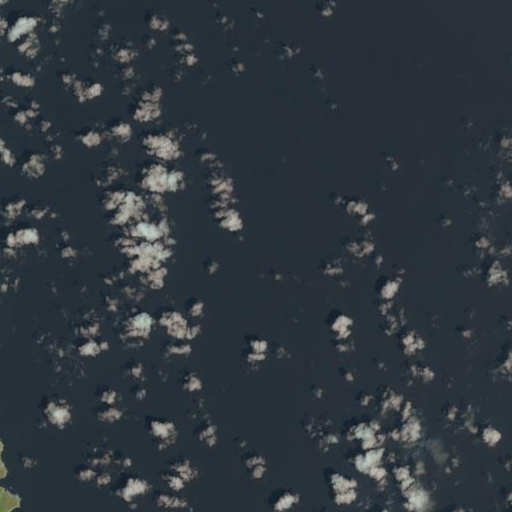}
		\includegraphics[width=0.24\textwidth]{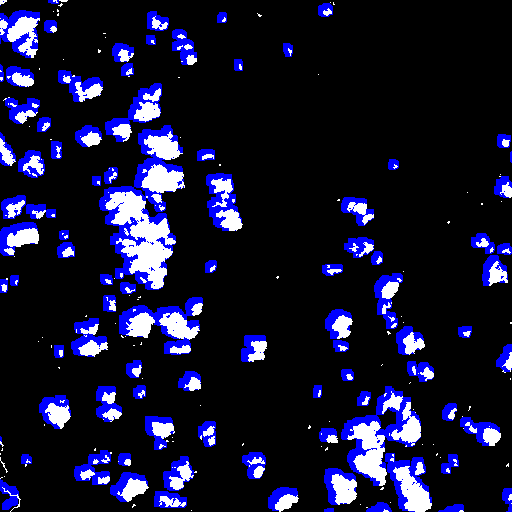} 
		\includegraphics[width=0.24\textwidth]{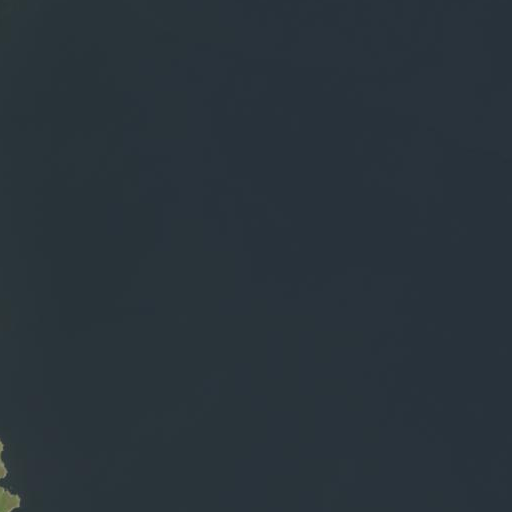}
		\\
		\vspace{2mm}
		\includegraphics[width=0.24\textwidth]{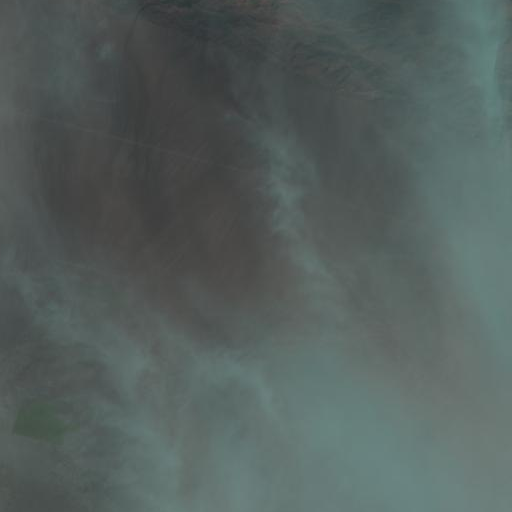}
		\includegraphics[width=0.24\textwidth]{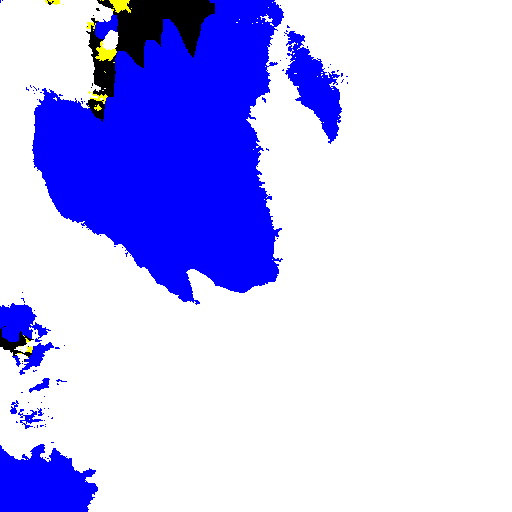} 
		\includegraphics[width=0.24\textwidth]{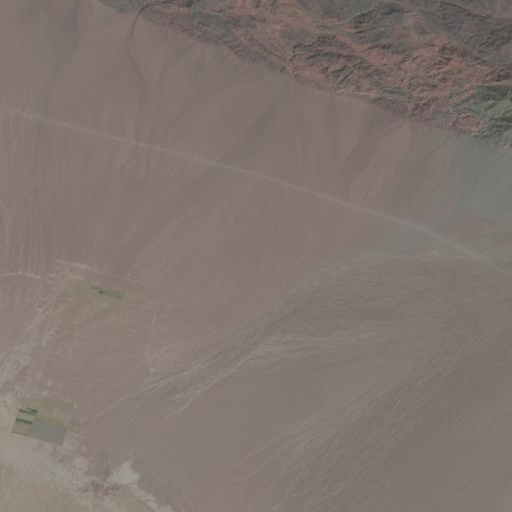}
		\\
		\vspace{2mm}
		\includegraphics[width=0.24\textwidth]{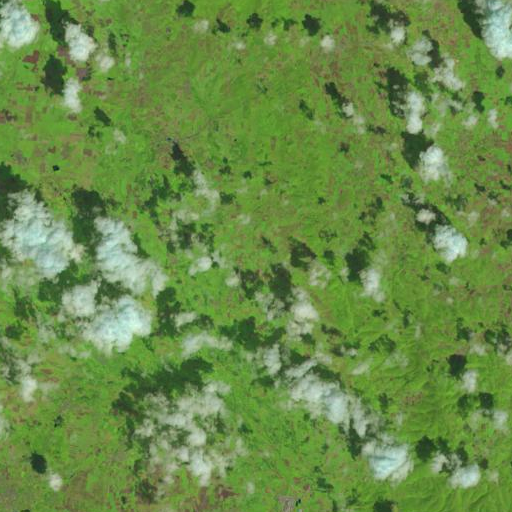}
		\includegraphics[width=0.24\textwidth]{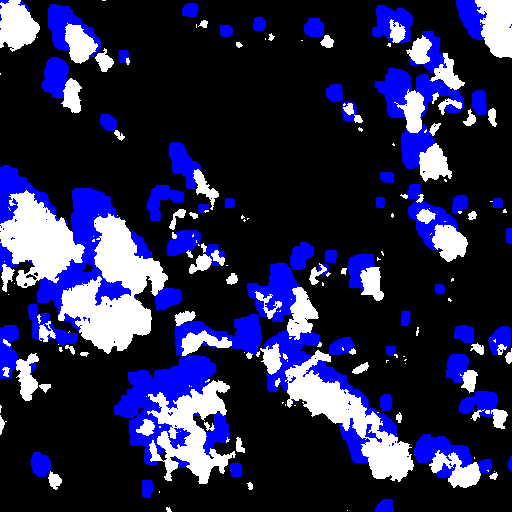} 
		\includegraphics[width=0.24\textwidth]{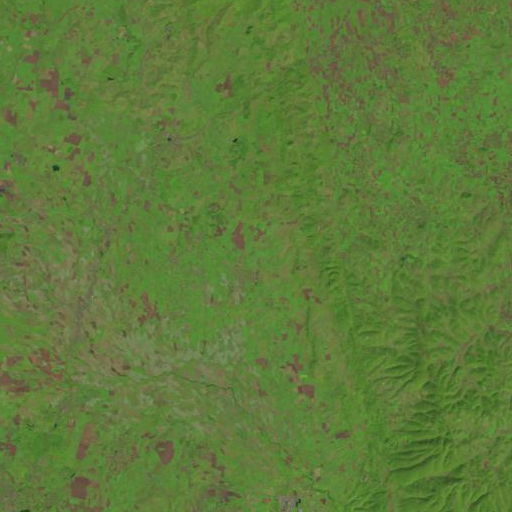}
		\\
		\vspace{2mm}
		\includegraphics[width=0.24\textwidth]{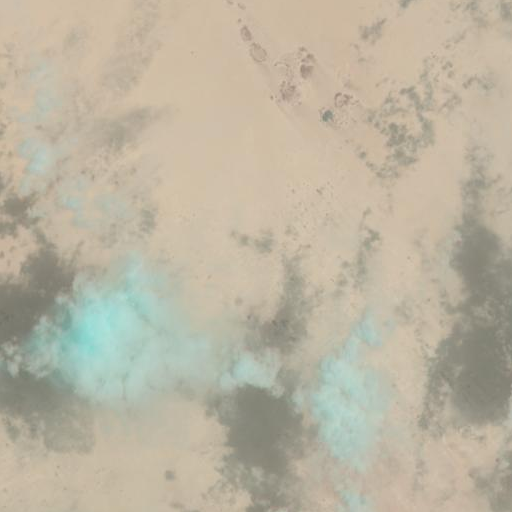}
		\includegraphics[width=0.24\textwidth]{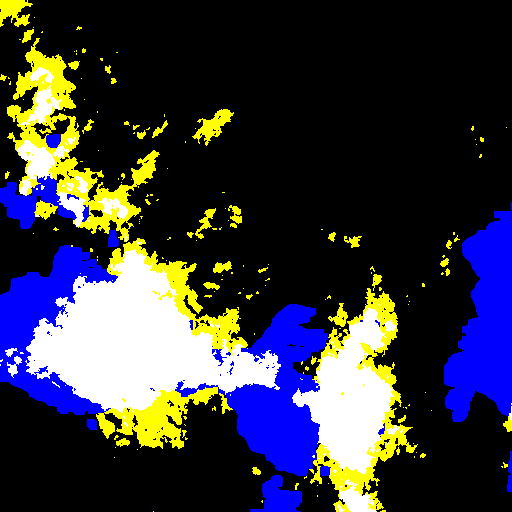} 
		\includegraphics[width=0.24\textwidth]{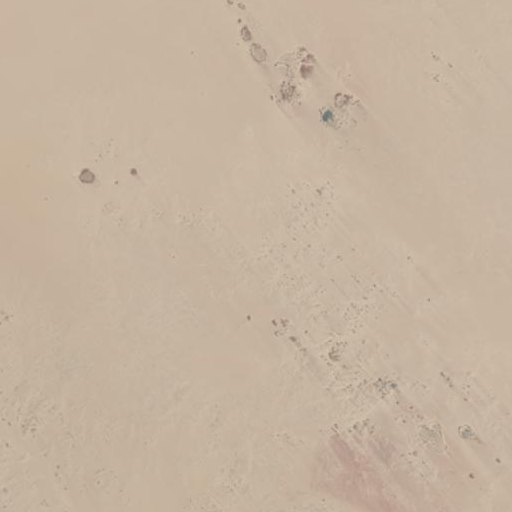}
	\end{center}
	\vspace{-5mm}
	\caption{RICE-\romannum[2] dataset.}
	\label{fig:rice2}
\end{figure}

In order to promote the development of deep learning in the field of remote sensing image de-clouding, we first proposed a benchmark dataset, named RICE (Remote sensing Image Cloud rEmoving), including RICE-\romannum[1] and RICE-\romannum[2], as shown in Figure~\ref{fig:rice1} and Figure~\ref{fig:rice2}, respectively.

\section{RICE Dataset}

\subsection{RICE-\uppercase\expandafter{\romannumeral1}}
RICE1 is collected on Google Earth, and the corresponding cloud and cloudless images are obtained by setting whether to display the cloud layer and then the acquired images are cut to 512*512 size without overlapping. RICE1 contains a total of 500 pairs of images. 

\subsection{RICE-\uppercase\expandafter{\romannumeral2}}
The RICE2 data is derived from the Landsat 8 OLI/TIRS dataset, and we use the LandsatLook Images with Geographic Reference in Earth Explorer. LandsatLook images are full-resolution files derived from Landsat Level-1 data products. LandsatLook images include Natural Color Image, Thermal Image and Quality Image. We used Natural Color Image and Quality Image in our dataset. The LandsatLook Natural Color image is a composite of Three bands (Landsat 8 OLI, Bands 6,5,4) to show a “natural” looking (false color) image.
LandsatLook Quality images are 8-bit files generated from the Landsat Level-1 Quality band to provide a quick view of the quality of the pixels within the scene to determine if a particular scene would work best for the user's application. Color mapping assignments can be Seen in the Table~\ref{tab:QuaImh}.

\begin{table}[h]
	\centering
	\caption{LandsatLook quality images designations.}
	\label{tab:QuaImh}
	\begin{tabular}{|l|l|l|l|l|}
		\hline
		\textbf{Description} & Designated Fill & Cloud & Cloud Shadow & Cirrus \\ \hline
		\textbf{Color} & Black & White & Blue & Yellow \\ \hline
	\end{tabular}
\end{table}

In order to get a cloudless reference image, we manually selected a cloudless image at the same location with a cloud image time less than 15 days apart. 

\section{Experiments}

We used the pix2pix~\cite{pix2pix2016} method to experiment on the dataset. For the RICE-\romannum[2] dataset, the input of pix2pix is a concation with a cloud image and a mask. We use PSNR and SSIM to measure the accuracy of the test data. The experimental results are shown in the Table~\ref{tab:psnr}.

\begin{table}[h]
	\centering
	\caption{LandsatLook quality images designations.}
	\label{tab:psnr}
	\begin{tabular}{|l|l|l|}
		\hline
		\textbf{Dataset} & PSNR & SSIM \\ \hline
		\textbf{RICE-\romannum[1]} & 31.03 & 0.91  \\ \hline
		\textbf{RICE-\romannum[2]} & 30.04 & 0.80 \\ \hline
	\end{tabular}
\end{table}

\section{Conclusion}

This paper introduced RICE, a Remote sensing Image Cloud rEmoving dataset,  to promote the development of deep learning in remote sensing image cloud removal tasks.  In the future, we will expand the RICE dataset, such as adding different resolutions and more scene categories.

\bibliographystyle{IEEEbib}
\bibliography{igass}

\end{document}